\documentclass[10pt,twocolumn,letterpaper]{article}

\pdfoutput=1

\usepackage{times}
\usepackage{graphicx}
\usepackage{amsmath}
\usepackage{amssymb}

\begin{document}

\title{Siamese Network of Deep Fisher-Vector Descriptors for Image Retrieval}

\author{Eng-Jon Ong, Sameed Husain and Miroslaw Bober\\
University of Surrey\\
Guildford, UK\\
{\tt\small e.ong,sameed.husain,m.bober@surrey.ac.uk}
}

\maketitle

\begin{abstract}
This paper addresses the problem of large scale image retrieval, with the aim of accurately ranking the similarity of a large number of images to a given query image.
To achieve this, we propose a novel Siamese network. This network consists of two computational strands, each comprising of a CNN component followed by a Fisher vector component. The CNN component produces dense, deep convolutional descriptors that are then aggregated by the Fisher Vector method. Crucially, we propose to simultaneously learn both the CNN filter weights and Fisher Vector model parameters. This allows us to account for the evolving distribution of deep descriptors over the course of the learning process. We show that the proposed approach gives significant improvements over the state-of-the-art methods on the Oxford and Paris image retrieval datasets. Additionally, we provide a baseline performance measure for both these datasets with the inclusion of 1 million distractors.
\end{abstract}

\section{Introduction}
The rise of digital cameras and smart phones, the standardization of computers and multimedia formats, the ubiquity of data storage devices and the technological maturity of network infrastructure has exponentially increased the volumes
of visual data available on-line and off-line. With this dramatic growth,
the need for an effective and computationally efficient content search system has become 
increasingly important. Given a large collection of images and videos, the aim is to retrieve individual images and video shots depicting instances of a user-specified object (query). 
There are a range of important applications for image retrieval including management of multimedia content, mobile commerce, surveillance, augmented automotive navigation etc. 
Performing robust and accurate visual search is challenging due to factors such as changing object viewpoints, scale, partial occlusions, varying backgrounds and imaging conditions.
Additionally, today's systems must be highly scalable to accommodate the the huge volumes of multimedia data, which can comprise billions of images. 

In order to overcome these challenges, a compact and discriminative image representation is required.
Typically, this is achieved by the aggregation of multiple local descriptors from an image into a single high-dimensional global descriptor.
The similarity of the visual content in two images is determined using a distance metric (e.g. Hamming or Euclidean distance) between their corresponding global descriptors. 
The retrieval is accomplished by calculating a ranking based on the distances between a set of images to a given query image.

This paper addresses the task of extracting a global descriptor by means of aggregating local deep descriptors. We achieve this using a novel combined CNN and Fisher Vector model that is learnt simultaneously. We also show our proposed model provides significant improvements in the retrieval accuracy when compared with related state-of-the-art approaches across different descriptor dimensionalities and datasets.

\begin{figure*}[t!]
\begin{center}
\includegraphics[width = 0.8\linewidth]{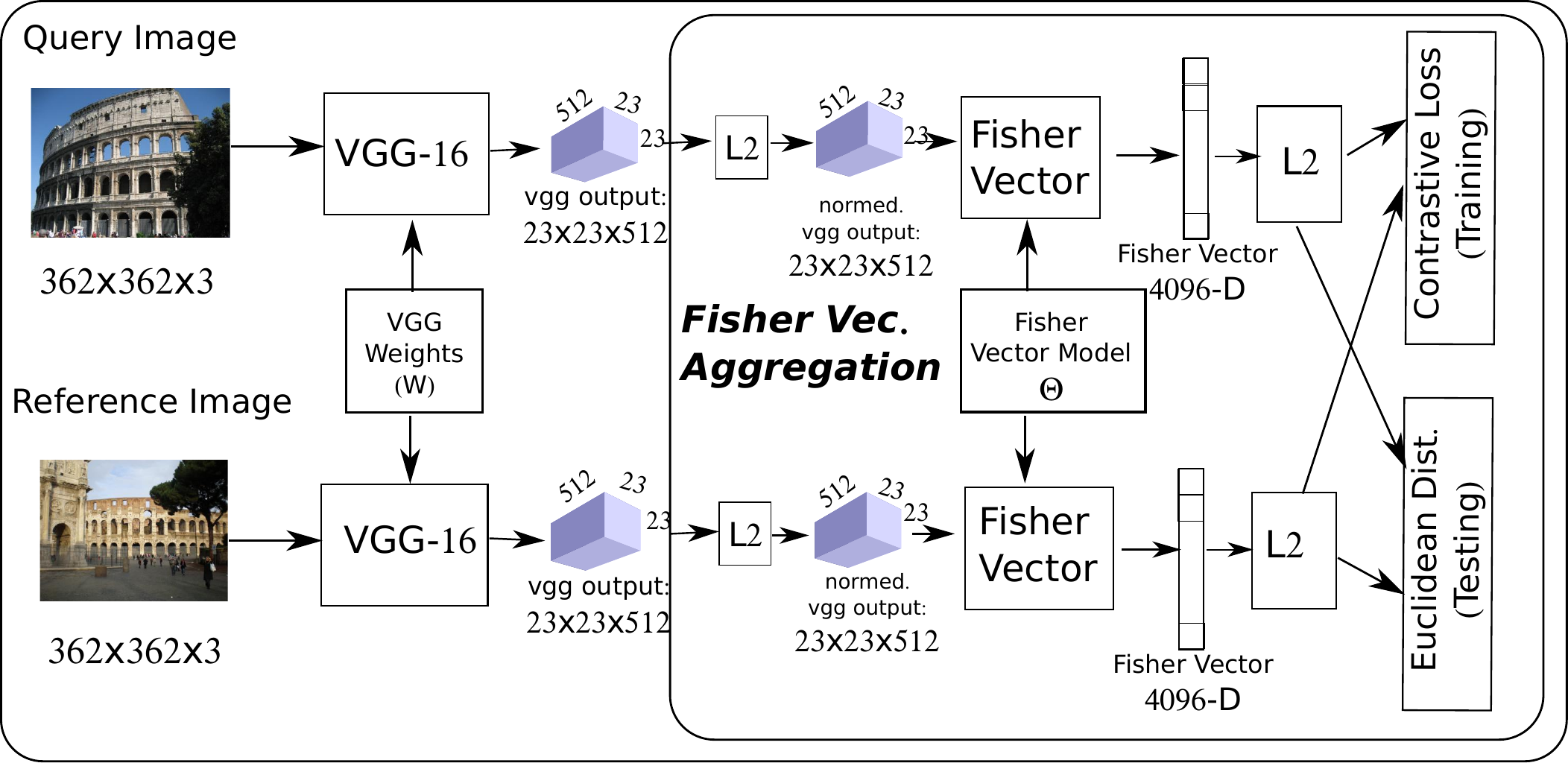} 
\end{center}
\caption{Overview of the training configuration of the proposed CNN-FV siamese network. During training, the last layer is the contrastive loss layer. During testing, the Euclidean distance between the final two fisher vectors is given. }
\label{fig:nn_overview}
\end{figure*}

\subsection{Related Work}
One popular method for generating global descriptors for image matching is the Fisher Vector (FV) method, which  aggregates local image descriptors (e.g. SIFT \cite{Lowe04}) based on the Fisher Kernel framework. A Gaussian Mixture Model (GMM) is used to model the distribution of local image descriptors, and the global descriptor for an image is obtained by computing and concatenating the gradients of the log-likelihoods with respect to the model parameters.
%
%
One advantage of the FV approach is its encoding of higher order statistics, resulting in a more discriminative representation and hence better performance \cite{PerronninD07}. 

The FV model is learnt using unsupervised clustering, and therefore cannot make use of matching and non-matching labels that are available in image retrieval tasks. One way of overcoming this shortcoming was proposed by Perronnin et al. \cite{PerronninL15}, where a fully connected neural network (NN) was trained by using the FV global descriptors as input.
Here, the fisher-vector model was initially learnt in an unsupervised fashion on extracted SIFT features. The FV model then produces input feature vectors for the fully connected NN, which in turn is learnt in a supervised manner using backpropagation. 

However, both the SIFT features and FV model in the above method are unsupervised. An alternative is to replace the low-level SIFT-features with deep convolutional descriptors obtained from convolutional neural networks (CNNs) trained on large-scale datasets such as ImageNet. 
Recent research has shown that image descriptors computed using deep CNNs achieve state-of-the-art performance for image retrieval and classification tasks. Babenko et al. \cite{BabenkoL15} aggregated deep convolutional descriptors to form global image representations: FV, Temb and SPoC. The SPoC signature is obtained by sum-pooling of the deep features. Razavian et al. \cite{RazavianSMC14} compute an image representation by the max pooling aggregation of the last convolutional layer. The retrieval performance was further improved when the RVD-W method was used for aggregation of CNN-based deep descriptors \cite{7577858}.


All of the above approaches use fixed pre-trained CNNs. However, these CNNs are trained for the purpose of image classification (e.g. 1000 classes of Imagenet) and may perform sub-optimally in the task of image retrieval. To tackle this, Radenovic et al. \cite{Radeno2016} and Gordo et al. \cite{Gordo2016} both proposed to use a Siamese CNN with max-pooling for aggregation. The CNN was fine-tuned on an image retrieval dataset. Two types of loss function were considered for optimisation: 1) the contrastive loss  function \cite{Radeno2016} and 2) the triplet loss function \cite{Gordo2016}. Both were able to achieve significant improvements from existing retrieval mAP scores. However, both these approaches use max-pooling as an aggregation method. The work proposed in this paper improves on this by employing a Fisher Vector model for aggregation instead of max-pooling. We also consider an alternative method of sum-pooling and compare different aggregation methods on standard benchmarks.

\subsection{Contributions and Overview}
The main contribution of this paper is a Siamese deep net that aggregates CNN-based local descriptors using the Fisher Vector model. Importantly, we propose to learn the parameters of the CNN and Fisher vectors simultaneously using stochastic gradient descent on the contrastive loss function. This allows us to adjust the Fisher vector model to account for changes in the distribution of the underlying CNN features as they are learnt on image retrieval datasets.
We also show that our proposed method improves on the retrieval performance of the following state-of-the-art approaches: Siamese CNN with max-pooling\cite{Radeno2016} and Triplet loss with max-pooling \cite{Gordo2016}. We show that our approach achieves mAP scores that equal or improve on state of the art results for the Oxford (81.5\%) and Paris datasets (82.5\%). Importantly, this was achieved without any segmentation of images used in \cite{Gordo2016}.
We also provide a new baseline of retrieval performance of our method when 1 million distractors are included into the test datasets.

The rest of the paper is structured as follows: Section \ref{sec:deep_fv} describes the proposed CNN-FV Siamese network used in this paper. The details for learning this network is given in Section \ref{sec:learn}. The experimental results are then described in Section \ref{sec:exp} before concluding in Section \ref{sec:conclusions}.

\section{Deep Fisher Vector Siamese Network}
\label{sec:deep_fv}
In this section, we describe the novel DNN that will learn a deep fisher vector representation by simultaneously learning the fisher-vector model components along with the underlying convolutional filter weights in a Siamese network.
The overview diagram of the proposed deep Siamese Fisher Vector network is shown in Fig. \ref{fig:nn_overview}. 

Traditionally, a Siamese network consists of two parallel branches in the network, where both branches share the same convolutional weights. One branch is fed a query image and the other branch a reference image which propagate through the network yielding 2 global descriptors respectively, which can be compared using Euclidean distance. Our proposed Siamese network is different in that each branch consists of two components: a CNN for producing deep image descriptors that are then aggregated via a Fisher Vector layer to produce the final global descriptor. 

\subsection{CNN-based Deep Descriptors}
Suppose the input image is given as $\mathbf{x} \in \mathcal{R}^{S\times S \times 3}$. In order to extract the deep convolutional descriptors from the CNN component, the input image is first passed through number of convolutional layers. Here, we use convolutional layers with the same structure as the VGG-16 \cite{VGG16} network with the fully connected layers removed. 

The CNN is effectively parameterised by the filter weights at each of its convolutional layers. We shall denote the collection of all the CNN filter weights as $W$. Formally, the CNN component can then be described by the function $f:\mathcal{R}^{S\times S \times 3} \rightarrow \mathcal{R}^{O\times O \times F}$, where $F$ is the final number of convolutional filters, each producing a convolutional image of size $O\times O$. We then treat the final layer as producing a set of $N_C = O\times O$ number of deep convolutional features that are of dimension $F$.

\subsection{Fisher Vectors}
In order to aggregate the $N_C$ deep convolutional features, we employ the method of Fisher Vectors. Firstly, let $\mathcal X=\{x_{t}\in \mathbb{R}^{d}, t=1...T \} $ be the set of $N_C$ $F$-dimensional deep convolutional features extracted from an image $I$. Let $u_{\Theta}$ be an image-independent probability density function which models the generative process of $\mathcal X$, where $\Theta$ represents the parameters of $u_{\Theta}$. 

A Gaussian Mixture model (GMM) \cite{Perronnin10}, $u_{\Theta}$ is used to model the distribution of the convolutional features, where: 
\[
u_{\Theta}(x)=\sum\limits_{j=1}^C \omega_{j} u_{j}(x)
\]

We represent the parameters of the $C$-component GMM by $\Theta =(\omega_{j}, \mu_{j}, \Sigma_{j}:j=1,...,C),$ where $\omega_{j}, \mu_{j}, \Sigma _{j}$ are respectively the weight, mean vector and covariance matrix of Gaussian $j$. The covariance matrix of each GMM component $j$ is assumed to be diagonal and is denoted by $\sigma_{j}^{2}$. The GMM assigns each descriptor $x_{t}$ to Gaussian $j$ with the soft assignment weight ($\tau_{tj}$) given by the posteriori probability:
\begin{eqnarray}
	\tau_{tj}=\frac{{exp}(-\frac{1}{2}{(\mathbf{x}_{t}-\boldsymbol{\mu} 
			_{j})}^{T}{\Sigma}_{j}^{-1}(\mathbf{x}_{t}-\boldsymbol{\mu}_{j}))}{{\sum\limits_{i=1}^n 
			{exp}(-\frac{1}{2}{(\mathbf{x}_{t}-\boldsymbol{\mu}_{i})}^{T}{\Sigma 
			}_{i}^{-1}(\mathbf{x}_{t}-\boldsymbol{\mu}_{i}))} }
	\label{eqn: soft assignment w}
\end{eqnarray}

The GMM can be interpreted as a probabilistic visual vocabulary, where each Gaussian forms a visual word or cluster.
The $d$-dimensional derivative with respect to the mean $\boldsymbol{\mu}_{j}$ of Gaussian $j$ is denoted by $\boldsymbol{\zeta}_{j}$:
\begin{eqnarray}
	\boldsymbol{\zeta}_{j}=\frac{1}{T\sqrt \omega_{j} }\sum\limits_{t=1}^T \tau_{tj} 
	\Sigma^{-1}_{j}(\mathbf{x}_{t}-\boldsymbol{\mu}_{j})
	\label{eqn: gradient with respect to mean}
\end{eqnarray}

We denote the elements of of $\boldsymbol{\zeta}_j$ as $\zeta_{jk}, k \in \{1,...,d\}$.
The final FV representation used, $\boldsymbol{\zeta}$, of image $I$ is obtained by concatenating the gradients $\boldsymbol{\zeta}_{j}$  for all Gaussians $j=1..n$ and normalising, giving: $\boldsymbol{\hat{\zeta}} = (\hat{\zeta}_{1,1}, \hat{\zeta}_{1,2},..., \hat{\zeta}_{C,d})$, with $ \hat{\zeta}_{jk} = \zeta_{jk}/|\boldsymbol{\zeta}|$, where
$|\boldsymbol{\zeta}| = \sum^{C,d}_{j,k}\sqrt{\zeta_{jk}^2}$. The dimensionality of $\boldsymbol{\zeta}$ is $d\times C$. Since the FV $\boldsymbol{\zeta}$ will be integrated into a Siamese-CNN, we shall henceforth refer to $\boldsymbol{\zeta}$ as ``SIAM-FV'' for \textbf{SIAM}ese-CNN-based \textbf{F}isher \textbf{V}ector.

\subsection{Fisher Vector Partial Derivatives}
\label{sec:fv_deriv}
In this section, the partial derivatives of the Fisher vector $\boldsymbol{\zeta}$ with respect to its underlying parameters ($\Theta$) are given. These partial derivatives will be used for learning the proposed deep net.
Firstly, we give the partial derivatives for the element ($\zeta_{jk}$) of $\boldsymbol{\zeta}_j$ for some cluster $j \in \{1,...,C\}$ and dimension, $k \in \{1,...,d\}$:
\begin{eqnarray}
\frac{\partial \zeta_{jk}}{\partial \omega_j} & = & 
-\frac{1}{2T(\omega_j)^{3/2}}\sum^{T}_{t=1}\frac{\tau_{tj}(x_{tk} - \mu_{jk})}{\sigma_{jk}} 
\label{eq:zeta_omega_grad}
\\
\frac{\partial \zeta_{jk}}{\partial \sigma_{jk}} & = & 
\frac{1}{T\sqrt{\omega_j}} \sum^T_{t=1}(x_{tk} - \mu_{jk})\left[ \frac{\sigma_{jk}\frac{\partial \tau_{tk}}{\partial \sigma_{jk}} - \tau_{tk} }{\sigma_{jk}^2} \right]
\label{eq:zeta_sigma_grad}
\\
\frac{\partial \zeta_{jk}}{\partial \mu_{jk}} & = & 
\frac{1}{T\sqrt{\omega_j}}
\sum^T_{t=1}
    \frac{
        \left[ 
        (x_{tk} - \mu_{jk})\frac{\partial \tau_{tj}}{\partial \mu_{jk}} - \tau_{tj}
        \right]
        }{\sigma_{jk}}
\label{eq:zeta_mu_grad}        
\\
\frac{\partial \zeta_{jk}}{\partial x_{tk}} & = &
\frac{1}{T\sigma_{jk}\sqrt{\omega_j}}
        \left[ 
        (x_{tk} - \mu_{jk})\frac{\partial \tau_{tj}}{\partial x_{tk}} + \tau_{tj}
        \right]
\label{eq:zeta_x_grad}
\end{eqnarray}

The partial derivatives of $\tau_{tj}$ in the above equations are detailed in Appendix \ref{app:tau_pd}. The equations Eq. \ref{eq:zeta_omega_grad} - \ref{eq:zeta_mu_grad} are used for calculating the gradients of the cluster prior, cluster mean and cluster standard deviation in the FV model. Eq. \ref{eq:zeta_x_grad} is used to backpropagate errors to the filter weights in the CNN component. We find that the partial derivatives of the final normalised fisher vector elements $\hat{\zeta}_{jk}$ all have the following form:
\begin{equation}
\frac{\partial \hat{\zeta}_{jk}}{ \partial \phi} = \frac{1}{|\boldsymbol{\zeta}|}\frac{\partial \zeta_{jk}}{\partial \phi} - \frac{\zeta_{jk}}{|\boldsymbol{\zeta}|^3}\sum^C_{j=1}\zeta_{jk}\frac{\partial \zeta_{jk}}{\partial \phi}
\label{eq:final_norm_pd}
\end{equation}
In order to obtain the exact partial derivative of $\hat{\zeta}_{jk}$ with respect to a particular parameter, we substitute $\phi$ with this parameter, look up the corresponding equation in Eq. \ref{eq:zeta_omega_grad}-\ref{eq:zeta_x_grad}, and substitute it into Eq. \ref{eq:final_norm_pd} above.

\section{Deep Learning of Fisher Vector Parameters}
\label{sec:learn}
It is possible to learn the Fisher Vector GMM parameters using the EM algorithm on the deep convolutional features. 
However, this is an unsupervised method that does not make use of available labelling information. In order to tackle this shortcoming, we propose performing {\em supervised} learning of the GMM parameters. To this end, we treat the learning of the GMM parameters as part of learning process of a DNN. 

For the purpose of learning, we are given a training dataset of $T$ pairs of images, each image with resolution $S \times S$. Each pair of training images is associated with a label, where 1 denotes matching images and 0 denotes non-matching images. We denote the training dataset as: $\{((X_i, X'_i), Y_i)_{i=1}^T\}$, where $X_i, X'_i \in \mathcal{R}^{S\times S}$ and $Y_i \in \{0,1\}$. The value of the labels of $Y_i$ is 0 for matching examples and 1 for non-matching examples.

Next, we describe the contrastive loss \cite{Hadsell2006} used for learning the proposed FV-CNN network. Firstly, Euclidean distance is used to measure the difference between two Fisher vectors: $D(\boldsymbol{\zeta}, \boldsymbol{\zeta}') = ||\boldsymbol{\zeta} - \boldsymbol{\zeta}' ||$.

Now, let the CNN weights be $W$ and the set of all the Fisher Vector parameters $\Omega$. The loss function is defined as:
\begin{eqnarray}
L(W, \Omega, Y_i, (\zeta_i, \zeta'_i))  =  \frac{1}{2}Y_i(D(\zeta_i, \zeta_i'))^2 + \nonumber\\ 
 \frac{1}{2}(1-Y_i)( \max(0, \beta - D(\zeta_i, \zeta_i')))^2
\end{eqnarray}
where $\beta$ is the heuristically determined margin parameter.

In order to optimise the GMM and cluster weight parameters of the Fisher vector, $\Phi$, the partial derivatives of $L$ with respect to these respective parameters: $\partial L/ \partial \phi, \forall \phi \in \Theta$ are used. For conciseness, we will not write the arguments $(\boldsymbol{\hat{\zeta}}, \boldsymbol{\hat{\zeta}}')$ when referring to the distance function $D$. So, using the chain rule on $L$ gives:
\begin{equation}
\frac{\partial L}{\partial \phi}
 = \underbrace{ \big[YD - (1-Y)\max(0, \beta - D)\delta_{\beta - D > 0}\big]}_{\partial L/\partial D}\frac{\partial D}{\partial \phi}
\label{eq:part_L_part_phi}
\end{equation}

The first backpropagated partial derivative $\partial L / \partial D$ determines the amount of error present in the Fisher vectors of matching or non-matching pairs. The partial derivatives $\partial D/\partial \phi$ allows us to adjust the FV model parameters and can similarly be derived using the chain rule, giving:
\begin{eqnarray}
\frac{\partial D}{\partial \phi} & = & \sum^D_{i=1} 2(\hat{\zeta}_i - \hat{\zeta}'_i)
\left( 
\frac{\partial \hat{\zeta}_i}{\partial \phi} - \frac{\partial \hat{\zeta}'_i}{\partial \phi}
\right) \nonumber \\
& = & \sum^C_{j=1}\sum^d_{k=1} 2(\hat{\zeta}_{jk} - \hat{\zeta}'_{jk})
\left( 
\frac{\partial \hat{\zeta}_{jk}}{\partial \phi} - \frac{\partial \hat{\zeta}'_{jk}}{\partial \phi}
\right)
\label{eq:part_d_part_phi}
\end{eqnarray}
where $\boldsymbol{\zeta}$ and $\boldsymbol{\zeta}'$ are the 2 input Fisher vectors to the distance function $D$ and the partial derivatives of $\hat{\zeta}_{jk}$ and $\hat{\zeta'}_{jk}$ detailed in Section \ref{sec:fv_deriv}. 

The parameters are then updated by adding the present value to their respective partial derivatives multiplied by the learning rate $\alpha$: $\phi_{t+1} \leftarrow \phi_t \alpha\partial L/\partial \phi$.

\section*{Updating CNN Weights}
The updating of the CNN weights $W$ is performed in a similar manner to the standard backpropagation, with the following difference: The gradients backpropagated from the contrastive loss and fisher layer is given by: $\partial L/\partial D \times \partial D/\partial x_{tk}$ from Eq. \ref{eq:part_L_part_phi} and \ref{eq:part_d_part_phi}, with the partial derivatives $\partial \zeta_{jk}/\partial x_{tk}$ (Eq. \ref{eq:zeta_x_grad}) inserted in place of $\partial \zeta_{jk}/\partial \phi$.
Since the CNN part is located below the Fisher vector layer, the above Fisher Vector gradients will then be propagated downwards to update the CNN weights $W$. 

\section{Experiments}
\label{sec:exp}
For our experiments, the siamese network was learned on the Landmarks dataset used in \cite{Radeno2016}. Testing was performed on two independent datasets: Paris \cite{Philbin08} and Oxford Buildings \cite{Philbin07} with the mean average precision score reported. To test large scale retrieval, these datasets are combined with 1 million Flickr images \cite{Bober13}, forming the Oxford1M and Paris1M dataset respectively.
We followed the standard evaluation procedure and crop the query images of Oxford and Paris dataset, with the provided bounding box. The PCA transformation matrix is trained on the independent dataset to remove any bias.

\subsection{Network Details}
For the CNN component, the convolutional layers and respective filter weights of the VGG-16 network \cite{VGG16} was used. However, the max-pooling and ReLU layer at the final convolution layer was removed. 8 clusters was used for the Fisher vector GMM model, with their parameters initialised using the EM algorithm. This resulted in FV of dimensionality 4096. For retrieval purposes, we then perform PCA or LDA and whitening on the 4096-D Fisher vector, reducing dimensionalities to: 128D, 256D and 512D. In order to learn the PCA or LDA model, when the Oxford Buildings dataset is tested, the Paris dataset is used to build the PCA/LDA model, and vice versa.
The contrastive loss margin parameter was set to $\beta = 0.8$. We set the learning rate equal to 0.001, weight decay 0.0005 and momentum 0.5. Training is performed to at most 30 epochs.  
\begin{figure*}
\begin{center}
\begin{tabular}{ccc}
\includegraphics[width = 0.33\linewidth]{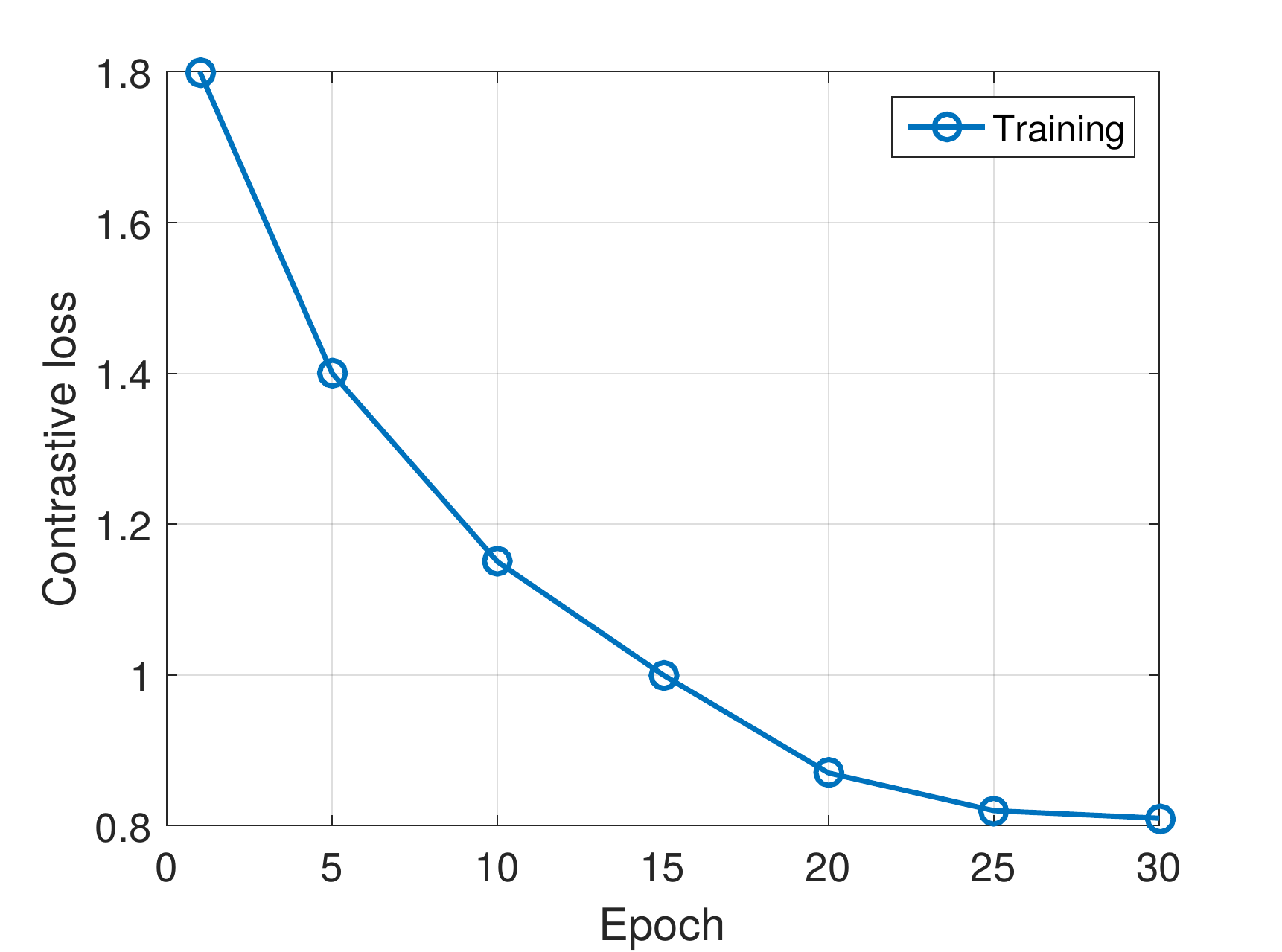} 
&\includegraphics[width = 0.33\linewidth]{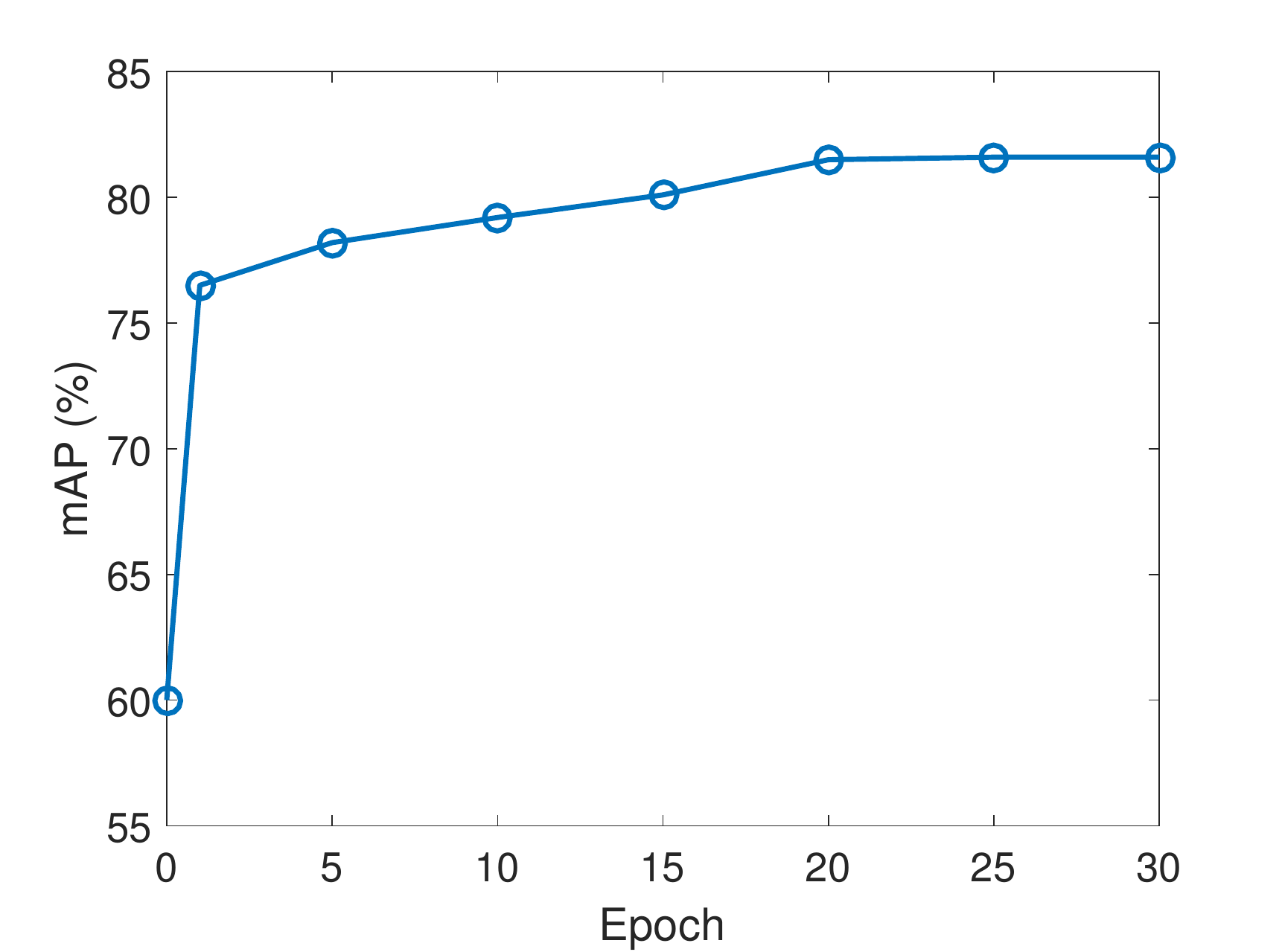} 
&\includegraphics[width = 0.33\linewidth]{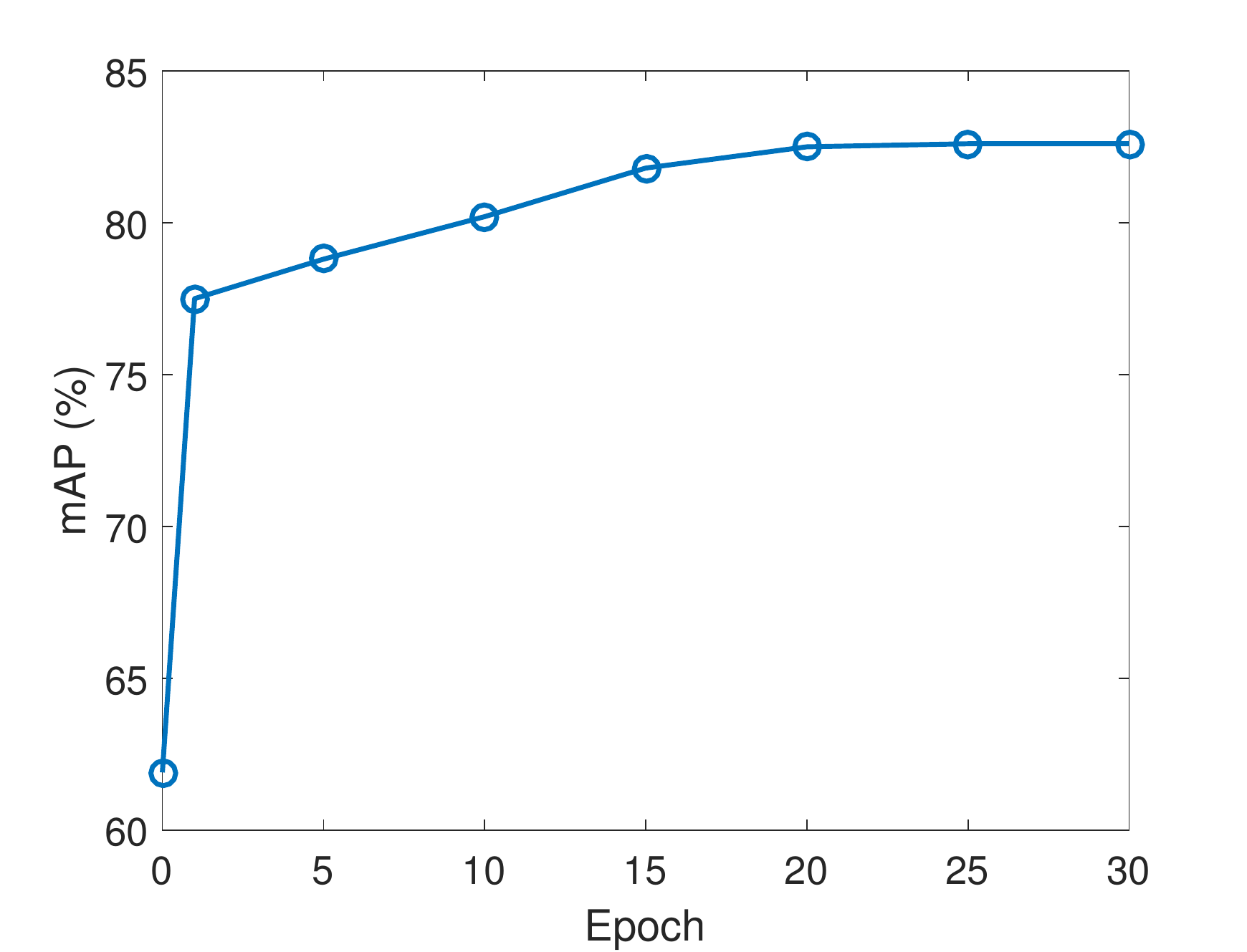}  \\
(a) Contrastive Loss & (b) mAP of Oxford Dataset & (c) mAP of Paris dataset.
\end{tabular}
\end{center}
\caption{a) Contrastive loss value during the training of the proposed deep Siamese Fisher-Vector network across different epochs. b,c) show the mAP scores for the Oxford (b) and Paris (c) datasets across different training epochs. }
\label{fig:contrastive_loss}
\end{figure*}
\begin{figure}
\begin{center}
\includegraphics[width = 0.98\linewidth]{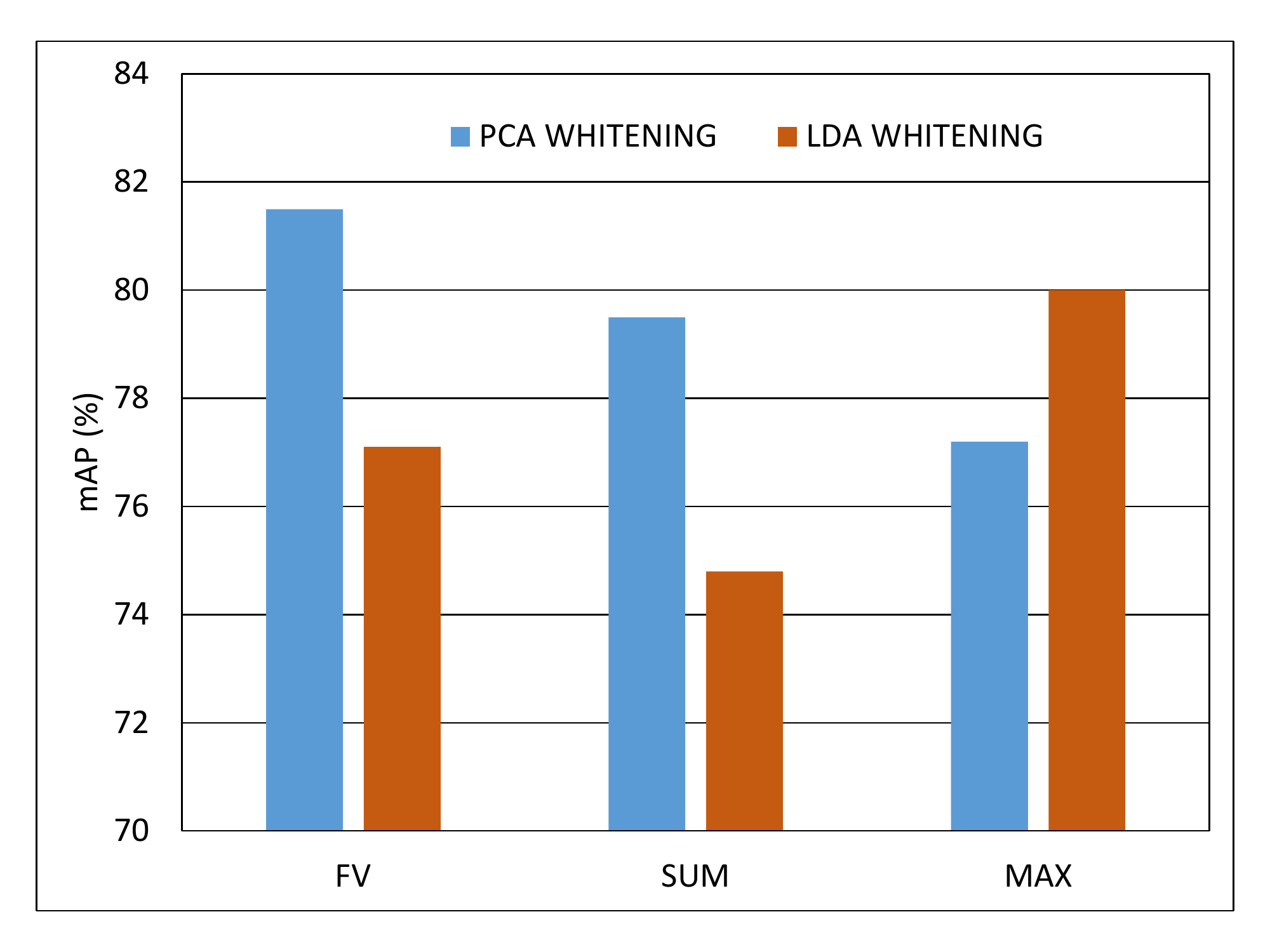} 
\end{center}
\caption{Shown are the mAP scores two dimensionality reduction methods: PCA and LDA on the proposed method (FV), sum-pooling and max-pooling \cite{Radeno2016} for the Oxford dataset.  }
\label{fig:pca_vs_lda}
\end{figure}
\subsection{Mining Non-Matching Examples}
There exists significantly more non-matching pairs compared with matching pairs. Therefore, exhaustive use of non-matching pairs will create a large imbalance in the number of matching and non-matching pairs used for training. In order to tackle this, only a subset of non-matching examples are selected via mining, which allows the selection of only ``hard'' examples used in \cite{Radeno2016}. Here, for each matching pair of images used, 5 of the closest non-matching examples to the query are used to form the non-matching pairs.

In this paper, 2000 matching pairs from the Landmarks dataset are randomly chosen. For each matching pair, 5 closest non-matching examples are then chosen, forming a 5-tuple, consisting of the following: query example; matching example; 5 non-matching examples. This forms a training set of $2000 + 5\times2000 = 12000$ pairs. This set of 12K pairs will be re-mined after 2000 iterations. In total, each epoch in the training cycle consists of 6000 iterations.

\subsection{Results}
In this section, we evaluate the different components of our system in terms of: retrieval performance of the SIAM-FV descriptor across different epochs; projection methods (PCA vs LDA); dimensionality of the SIAM-FV descriptor; and compare the performance to the latest state-of-the-art algorithms.

\begin{figure*}
\begin{center}
\begin{tabular}{cc}
\includegraphics[width = 0.5\linewidth]{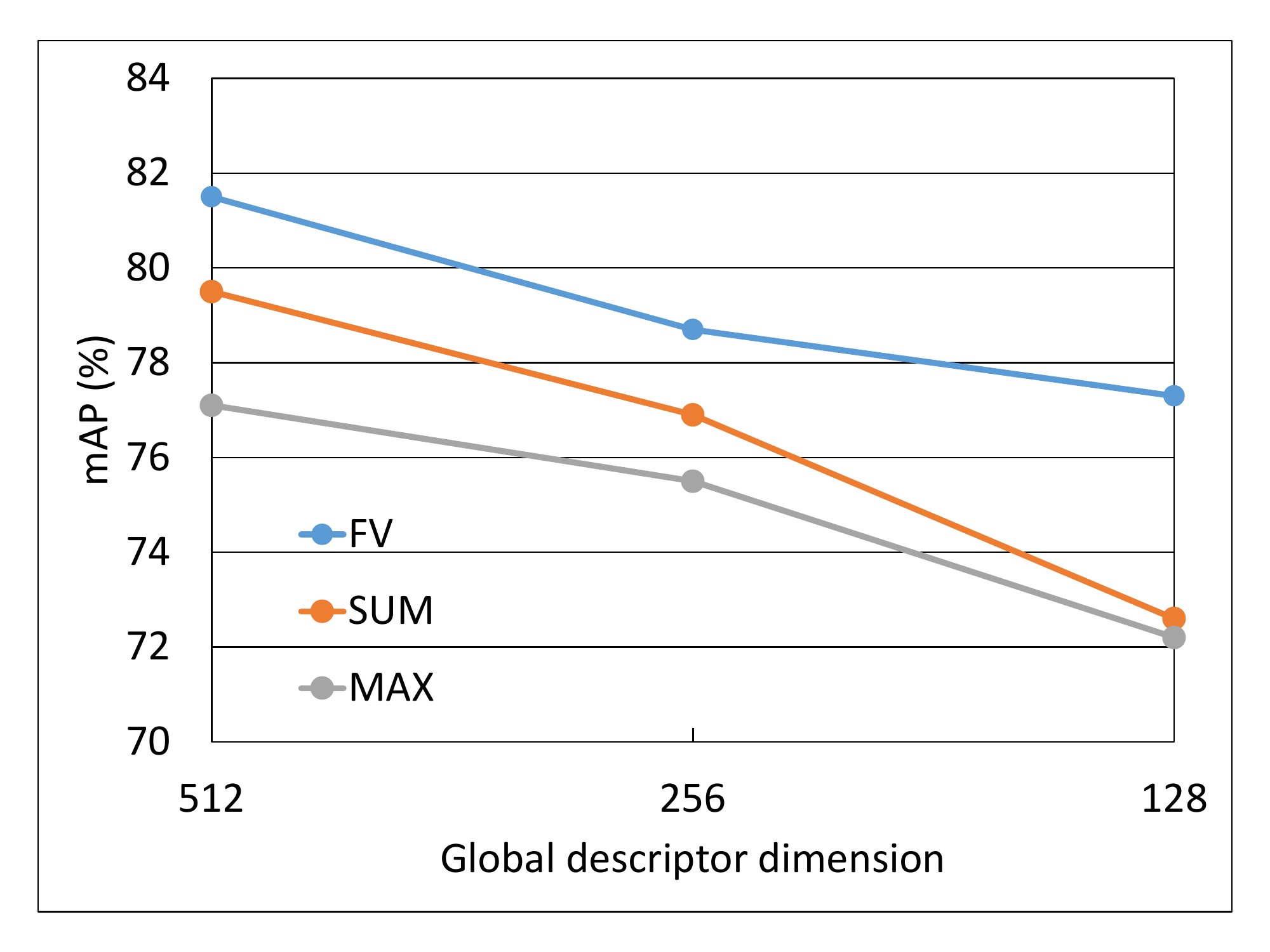} 
&\includegraphics[width = 0.5\linewidth]{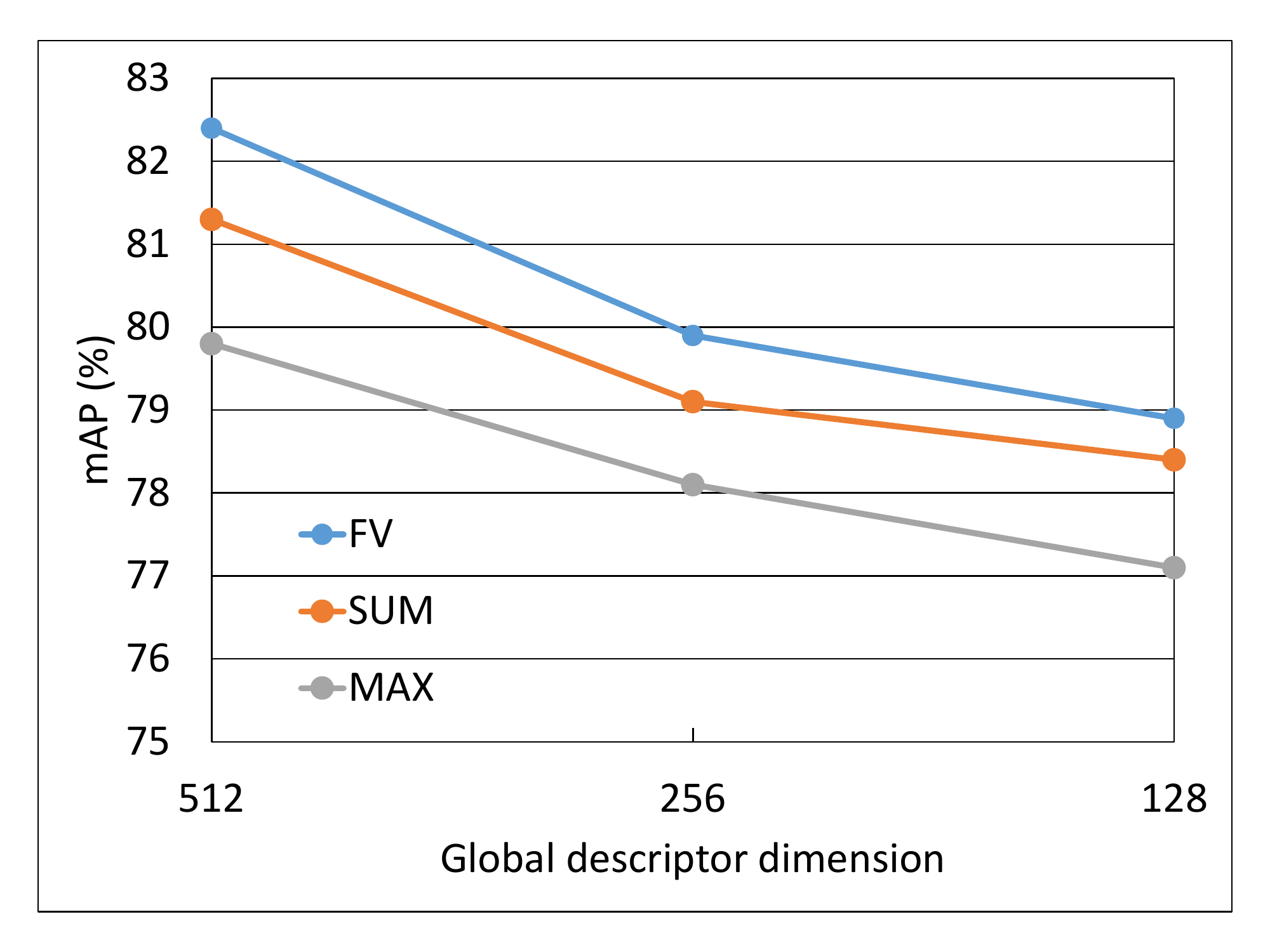}  \\
(a) Oxford & (b) Paris
\end{tabular}
\end{center}
\caption{Shown are the mAP scores for the different global descriptor methods: proposed FV global descriptor (FV), sum-pooling (SUM) and max-pooling (MAX) \cite{Radeno2016}, across different PCA-reduced dimensions and datasets: (a) Oxford; (b) Paris. }
\label{fig:dim_comp}
\end{figure*}
\subsection*{Learning}
\begin{figure*}
\begin{center}
\includegraphics[width = 0.95\linewidth]{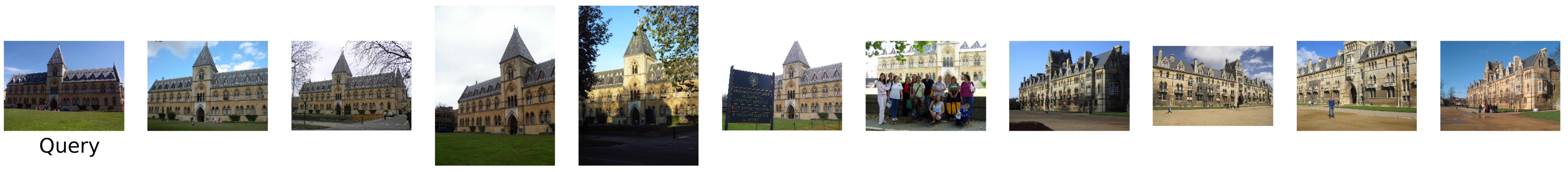} \\
{\tiny Average Precision: 100.00\%} \\
\includegraphics[width = 0.95\linewidth]{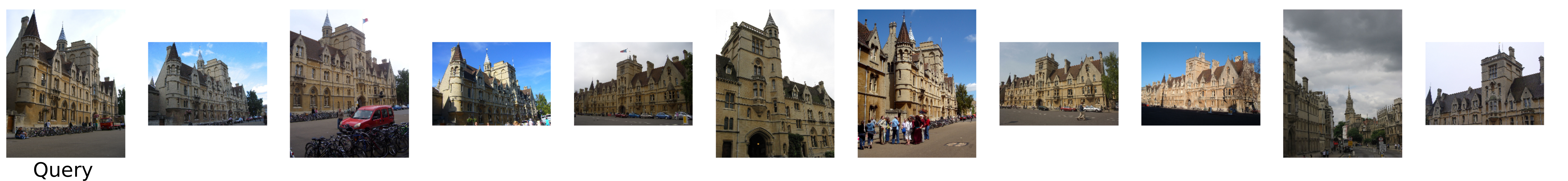} 
\\
{\tiny Average Precision: 87.98\%} \\
\includegraphics[width = 0.95\linewidth]{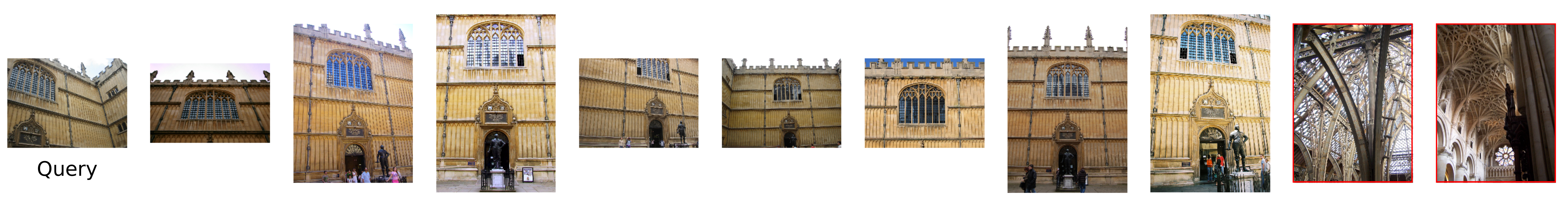} 
\\
{\tiny Average Precision: 65.52\%} \\
\includegraphics[width = 0.95\linewidth]{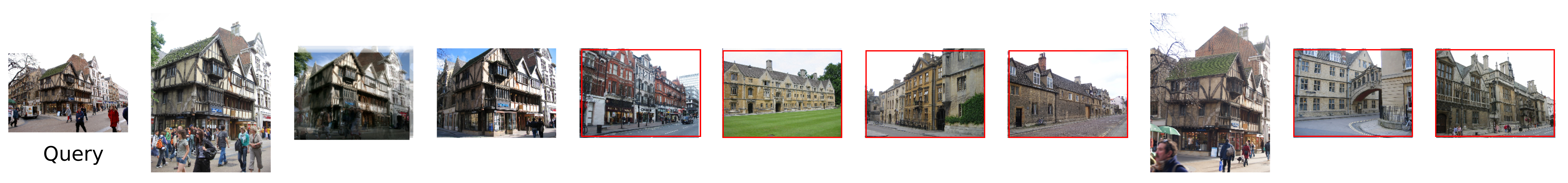} \\
{\tiny Average Precision: 45.11\%} \\
\end{center}
\caption{Examples of the top 10 retrieved images on the Oxford dataset using the proposed method for different average precisions. Retrieved non-matching images are highlighted in red.}
\label{fig:ret_imgs}
\end{figure*}

\begin{figure*}
\begin{center}
\includegraphics[width = 0.95\linewidth]{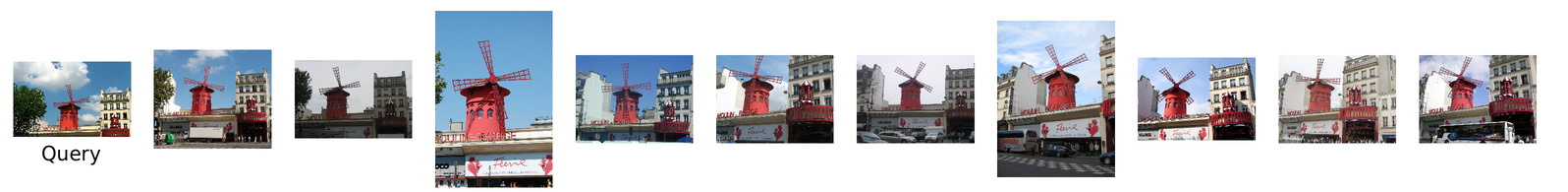} 
\\
{\tiny Average Precision: 97.21\%} \\
\includegraphics[width = 0.95\linewidth]{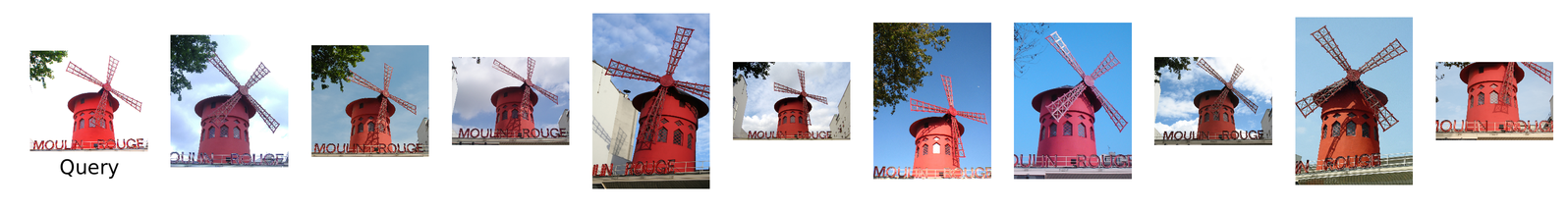} \\
{\tiny Average Precision: 85.30\%} \\
\includegraphics[width = 0.95\linewidth]{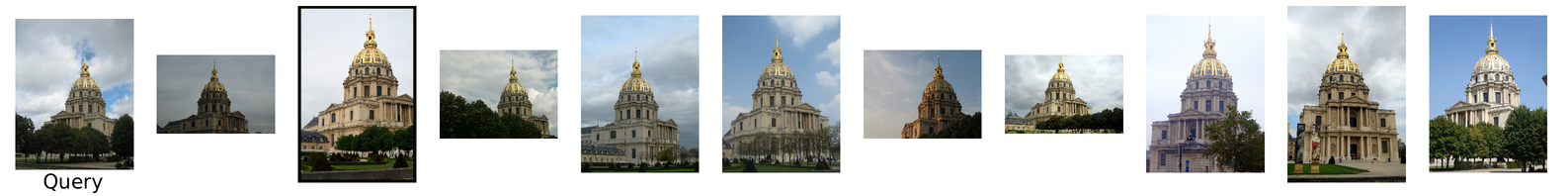} 
\\
{\tiny Average Precision: 78.88\%} \\
\includegraphics[width = 0.95\linewidth]{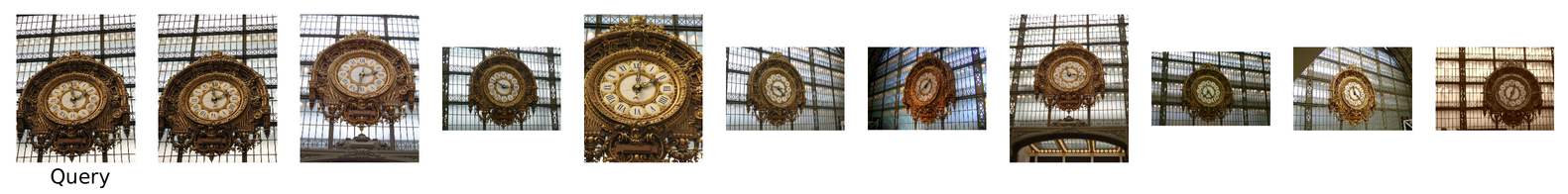} \\
{\tiny Average Precision: 38.39\%} \\
\end{center}
\caption{Examples of the top 10 retrieved images on the Paris dataset using the proposed method for different average precisions. }
\label{fig:ret_imgs2}
\end{figure*}

The behaviour of the contrastive loss during learning is shown in Fig.\ref{fig:contrastive_loss}. It can be seen here that the initial 20 epochs give a large reduction in the loss value, and subsequent epochs producing only small further improvements in the loss function. 
In terms of the mAP results on the test datasets of Oxford and Paris, we find that the greatest improvement is obtained from the initial 5 epochs, with approximately 14-16\% improvement in mAP scores. This can be seen in Fig. \ref{fig:contrastive_loss}b) for the Oxford dataset and Fig. \ref{fig:contrastive_loss}c for the Paris dataset. Examples of the retrieved images based on the SIAM-FV descriptor for the Oxford and Paris datasets can be seen in Fig. \ref{fig:ret_imgs} and \ref{fig:ret_imgs2} respectively.

\subsection*{Projection Methods: PCA vs LDA}
Fig. \ref{fig:pca_vs_lda} shows the mAP results achieved by employing PCA and LDA for dimensionality reduction  on the Oxford dataset. In \cite{Radeno2016}, it was found that for max-pooling aggregation, LDA provided better performance at 80.0\%, compared to PCA 76.1\%. However, the converse was found for our SIAM-FV descriptor, which achieves 81.5\% with PCA and 77.1\% using LDA. This was also found to be the case when sum-pooling was used, with 79.5\% for PCA vs 74.8\% using LDA. Thus for the remaining experiments, we have employed PCA as our choice for dimensionality reduction.

\subsection*{Dimensionality of SIAM-FV }
Figure. \ref{fig:dim_comp}a,b, demonstrates the performance of SIAM-FV signature when reduced to different dimensionalities via PCA+Whitening. As expected, the best performance is obtained when the dimensionality is highest, at 512D for both Oxford and Paris datasets. Crucially, the proposed SIAM-FV has a mAP score that is approximately 2\% higher than sum-pooling and 4\% higher than max-pooling on the Oxford dataset across all dimensionalities 128D,256D and 512D. This gain in performance is similar for the Paris dataset, with the SIAM-FV method outperforming both sum-pooling and max-pooling across all dimensionalities.

\subsection*{Comparison with State-of-the-Art}
This section compares the performance of the proposed method to the state-of-the-art algorithms. 
Table \ref{tab:res8k} summarises the results for medium footprint signatures (4k-512 dimensions). It can be seen that the proposed SIAM-FV representation outperforms most of the prior-art methods. On Paris dataset, the R-MAC representation provides marginally better performance. Note that R-MAC used region based pooling where deep features are max-pooled in several regions of an image using multi-scale grid. 

Gordo et al. \cite{Gordo2016} achieved 83.1\% on Oxford dataset. However they employed a region proposal network and extracted MAC signatures from 256 regions in an image, significantly increasing the extracting complexity of the representation. 

We now focus on a comparison of compact representations which are practicable in large-scale retrieval, as presented in Table \ref{tab:res128B}. The dimensionality of the SIAM-FV descriptor is reduced from 4096 to 128 via PCA. The results show that our method outperforms all presented methods. On the large dataset of Oxford1M SIAM-FV provides a gain of +2.4\% compared to latest MAC* signature.

\begin{table}[!t]
	\caption{Comparison with the state of the art using medium footprint signatures.}
	\label{tab:res8k}
	\centering
	\begin{tabular}{|l|c|c|c|c|c|c|}
		\hline
		Method & Size & Oxf5k & Oxf105k &Paris6k \\
		\hline \hline
	    TEmb \cite{Jegou14} &1024 &56.0 &50.2  &- \\ \hline 
	    NetVLAD \cite{Arandjelovic16} &4096 &71.6 &-  &79.7 \\ \hline
	    MAC \cite{Radeno2016} &512 &58.3 &49.2  &72.6 \\ \hline
	    R-MAC \cite{ToliasSJ15} &512 &66.9 &61.6  &\textbf{83.0} \\ \hline
	    CroW \cite{KalantidisMO15} &512 &68.2 &63.2  &79.7 \\ \hline
	    MAC* \cite{Radeno2016} &512 &80.0 &75.1  &82.9 \\
	    \hline
	    SUM Pool  &512 &79.5 &75.0  &81.3 \\\hline
	    SIAM-FV &512 &\textbf{81.5} &\textbf{76.6}  & 82.4  \\ \hline 
	\end{tabular}
\end{table}

\begin{table*}[!t]
	\caption{Comparison with the state of the art using small signatures.}
	\label{tab:res128B}
	\centering
	\begin{tabular}{|l|c|c|c|c|c|c|} \hline
     Method & Size & Oxf5k & Oxf105k &Oxf1M &Paris6k & Paris1M \\ \hline
    Max-pooling \cite{RazavianSMC14} &256  &53.3 &- &- &67.0 &-\\ \hline 
        SPoC \cite{BabenkoL15}  &256   & 53.1   &50.1   &- &- &-  \\  \hline
       MAC \cite{Radeno2016} &256 &56.9 &47.8  &- &72.4 &-   \\ \hline
        NetVLAD \cite{Arandjelovic16} &256 &63.5  &- &- &73.5 &-  \\ \hline
        CroW \cite{KalantidisMO15} &256 &65.4 &59.3 &- &77.9  &-\\ \hline
        Ng et al \cite{7301272}&128 &59.3 &- &- &59.0 &- \\ \hline
        MAC* \cite{Radeno2016} &128 &76.8 &70.8 &60.1 &78.8  &62.5 \\
	    \hline 
	    SUM Pool  &128 &72.6 &67.7 &57.9  &78.4 &62.4\\\hline 
	    SIAM-FV &128 &\textbf{77.3} &\textbf{71.8} & \textbf{62.5} &\textbf{78.9} & \textbf{63.2}\\\hline
	\end{tabular}
\end{table*}

\section{Conclusions}
\label{sec:conclusions}
In this paper, we have proposed a robust and discriminative image representation by aggregating deep descriptors using Fisher vectors. We have also proposed a novel learning method that allows us to simultaneously fine-tunes the deep descriptors and adapt the Fisher vector GMM model parameters accordingly. This effectively allows us to perform supervised learning of the Fisher vector model using matching and non-matching labels by optimising the contrastive loss. The result is a CNN-based Fisher vector (SIAM-FV) global descriptor. We have also found that PCA was a more suitable dimensionality reduction method compared with LDA when used with the SIAM-FV representation. 
We have shown that this model produces significant improvements in the retrieval mean average precision scores. On the large scale datasets, Oxford1M and Paris1M, SIAM-FV representation achieves a mAP of 62.5\% and 63.2\%, all yielding superior performance to the state-of-the-art.  

\appendix
\section{Partial Derivatives of $\tau_{tj}$}
\label{app:tau_pd}
We find that the partial derivatives of $\tau_{tj}$ with respect to $\sigma_{jk}$ and $\mu_{jk}$ both have the same form. So, let $\phi$ be either $\sigma_{jk}$,$\mu_{jk}$ or $x_{tk}$. Also, let the numerator of $\tau_{tj}$ be denoted as $\tau^{(j)}_{tj}$ and its denominator $\tau^{(\Sigma)}_{tj}$, so $\tau_{tj} = \tau^{(j)}_{tj}/\tau^{(\Sigma)}_{tj}$, then:
\begin{eqnarray*}
\frac{\partial \tau_{tj}}{\partial \phi} = 
\frac{
[\tau^{(\Sigma)}_{tj} - \tau^{(j)}_{tj}]\frac{\partial \tau^{(j)}_{tj}}{\partial \phi}
}
{(\tau^{(\Sigma)}_{tj})^2 },\;\; \phi \in \{\sigma_{jk},\mu_{jk}, x_{tk}\}
\end{eqnarray*}
where,
\begin{eqnarray*}
\frac{\partial \tau^{(j)}_{tj}}{\partial \mu_{jk}} & = & 
\tau^{(j)}_{tj}\left[ \frac{(x_{tk} - \mu_{jk})^2}{\sigma^3_{jk}} \right]\\
\frac{\partial \tau^{(j)}_{tj}}{\partial \sigma_{jk}} & = &
\tau^{(j)}_{tj}\left[ \frac{(x_{tk} - \mu_{jk})}{\sigma^2_{jk}} \right] \\
\frac{\partial \tau^{(j)}_{tj}}{\partial x_{tk}} & = &
\tau^{(j)}_{tj}\left[ \frac{\mu_{jk} - (x_{tk})}{\sigma^2_{jk}} \right] 
\end{eqnarray*}

{\small
\bibliographystyle{ieee}
}

\end{document}